\documentclass[conference]{IEEEtran}
\usepackage{amsmath,amssymb,amsfonts}
\usepackage{algorithmic}
\usepackage{graphicx}
\usepackage{subfigure}
\usepackage{textcomp}
\usepackage{xcolor}
\usepackage{multirow}
\usepackage{multicol}
\usepackage{array}
\usepackage{soul}
\usepackage{pdfpages} 
\usepackage{float}
\usepackage{todonotes}
\usepackage{acronym}

\usepackage{fancyhdr}
\usepackage[
backend=bibtex,
sorting=none,
style=ieee,
citestyle=numeric-comp
]{biblatex}

\bibliography{main}

\def\BibTeX{{\rm B\kern-.05em{\sc i\kern-.025em b}\kern-.08em
    T\kern-.1667em\lower.7ex\hbox{E}\kern-.125emX}}
\begin{document}

\title{
RX-ADS: Interpretable Anomaly Detection using Adversarial ML for Electric Vehicle CAN data
{\footnotesize }
\thanks{Identify applicable funding agency here. If none, delete this.}
}

\author{\IEEEauthorblockN{ Chathurika S. Wickramasinghe \textsuperscript{1}, Daniel L. Marino \textsuperscript{1}, Harindra S. Mavikumbure \textsuperscript{1}, Victor Cobilean\textsuperscript{1},\\ Timothy D. Pennington\textsuperscript{2}, Benny J. Varghese\textsuperscript{2}, Craig Rieger\textsuperscript{2}, Milos Manic\textsuperscript{1}}
	\IEEEauthorblockA{
	\textsuperscript{1} \textit{Virginia Commonwealth University},
	Richmond, VA, USA \\
  \textsuperscript{2} \textit{Idaho National Laboratory},
	Idaho Falls, Idaho, USA \\
	brahmanacsw@vcu.edu,
	marinodl@vcu.edu,  craig.rieger@inl.gov, misko@ieee.org}
}

 \maketitle
 
\thispagestyle{fancy}

\begin{abstract}
Recent year has brought considerable advancements in Electric Vehicles (EVs) and associated infrastructures/communications. Intrusion Detection Systems (IDS) are widely deployed for anomaly detection in such critical infrastructures. 
This paper presents an Interpretable Anomaly Detection System (RX-ADS) for intrusion detection in CAN protocol communication in EVs. 
Contributions include: 1) window based feature extraction method; 2) deep Autoencoder based anomaly detection method; and 3) adversarial machine learning based explanation generation methodology.
The presented approach was tested on two benchmark CAN datasets: OTIDS and Car Hacking.
The anomaly detection performance of RX-ADS was compared against the state-of-the-art approaches on these datasets: HIDS and GIDS.
The RX-ADS approach presented performance comparable to the HIDS approach (OTIDS dataset) and has outperformed HIDS and GIDS approaches (Car Hacking dataset).
Further, the proposed approach was able to generate explanations for detected abnormal behaviors arising from various intrusions. These explanations were later validated by information used by domain experts to detect anomalies. Other advantages of RX-ADS include: 1) the method can be trained on unlabeled data; 2) explanations help experts in understanding anomalies and root course analysis, and also help with AI model debugging and diagnostics, ultimately improving user trust in AI systems.


\end{abstract}


\begin{acronym}[ NY]  
\acro { IDS }       Intrusion Detection Systems 
\acro { ADS }        Anomaly Detection Systems 
\acro { ITS }        Intelligent Transportation Systems    
\acro { NN } Neural Networks 
\acro { XAI }        Explainable Artificial Intelligence 
\acro { CAN }        Controller Area Network 
\acro { HIDS }        Histogram-based IDS 
\acro { GIDS }        GAN-based IDS 
\acro { DoS }        Denial-of-Service
\acro { OTIDS }        Offset ratio and Time interval based IDS
\end{acronym}

\section{Introduction}


Electric Vehicles (EVs) are becoming a primary component in Intelligent Transportation Systems (ITSs) as it decreases fossil fuel consumption and greenhouse gas emissions, reducing negative environmental impact \cite{8807726}. 
In recent years, there has been a rapid growth in EV infrastructure, expanding to various areas, including EV manufacturing, charging stations, battery advancements, electric vehicle supply equipment, and other roadside infrastructures \cite{8344848, s18041212, Boukerche2019CrowdMT}. 
Within EV infrastructure, different communication technologies such as vehicle-to-Vehicle (V2V), Vehicle-to-sensor-board (V2S), vehicle-to-infrastructure (V2R), vehicle-to-human (V2H), and vehicle-to-internet (V2I) plays a major role in building resilient operations \cite{8377181}.
Security of these technologies is critical to avoid vulnerabilities such as DoS attacks, false data injections, spoofing and modification \cite{8377181, 8514157}. 

Intrusion Detection Systems (IDSs) are widely used approaches in critical infrastructures such as EV infrastructure \cite{ALOQAILY2019101842, 9243152}. 
The purpose of IDS is to detect attacks and intruders in communication systems of critical infrastructure, thus avoiding possible catastrophic failures and economic losses.
For example, in an EV, attacks can cause break malfunction, engine overheating, control steering issues, and door lock issues, resulting in life-threatening and catastrophic damages \cite{ALOQAILY2019101842}.
Not only EVs, other infrastructure components such as charging stations are prone to severe advanced persistent threats (APT) such as ransomware and malware \cite{9243152}.
Thus building IDSs has become a vital component within EV infrastructure. 

During the last decade, data-driven machine learning approaches such as Neural Networks (NNs) have been widely used for building IDSs for various critical infrastructure settings \cite{8591457}. 
There are two main type of IDSs: Signature based IDS and Anomaly Based IDS \cite{articleSIDSvsADS}. 
Typically, Anomaly Detection systems (ADSs) have the advantage of detecting both known attacks and unknowns/new attacks/abnormalities in the systems \cite{articleSIDSvsADS, KIM20141690}.
The idea of ADSs is to learn the normal behavior of a system such that anything outside learned normal behavior is detected as an anomaly.
The majority of ADSs are trained using only data coming from normal class/behavior. 
Therefore, it does not require expensive data labeling process (time-consuming, costly, and requires expertise in data) \cite{9373350}. 
Further, ADSs can be developed with an abundance of unlabelled data generated in real-world systems.
Out of widely used NN architectures for ADS development, Autoencoders (AE) has gained much attention. 
The reasons for this include many advantages of AEs such as in-build anomaly detection capability, can be trained with unlabelled data, scalability, feature extraction and dimentionality reduction capability. 
Therefore, in this paper, we are developing a AE based ADS. 

Trustworthy AI is a widely discussed topic when applying NNs for mission-critical infrastructures.
Despite the performance benefits of NNs, people hesitate to trust these systems. The main reason for this is the difficulty of understanding the decision-making process of the AI models, making these systems black-box models \cite{9142644}. 
It is crucial to address these trust-related issues to build trust between humans and these AI systems.
By addressing these issues, the Trustworthy AI research area has emerged. 
One main component of Trustworthy AI is the Explainability or Interpretability of AI systems (XAI). 
XAI aims to provide an understanding of black-box models, enabling users to question and challenge the outcomes of AI systems. 
It provides many advantages, including justifying outcomes of AI systems, improving trust on AI models, model debugging, and diagnosing \cite{9536751}.  
Therefore, this paper present an interpretable ADS developed using AEs.

This paper present the following contributions:
\begin{enumerate}
   \item Feature Extraction: Window based feature engineering approach which uses a overlapping sliding window of data frames to extract cyber features. 
    \item Anomaly Detection: ResNet AE based Anomaly Detection System Framework: Framework only used baseline behavior data for learning the normal behavior of the system, thus any deviation from the baseline are tagged as anomaly.
    \item Novel Explainable ADS: Explanations for anomalous behaviors by generating adversarial samples. These explanation helps with understanding anomalous behavior, understanding the decision making process of AE, and distinguishing different types of anomalies. 
\end{enumerate}

The presented approach was tested on two benchmark datasets which were provided by the Hacking and Countermeasures Research Laboratory. 
This approach was developed for an ongoing effort with Idaho National Laboratory (INL) to build ADS for an EV charging system (EVCS). 
Specifically for EVCSs, RX-ADS can provide multiple advantages, including understanding the root courses of a given anomaly, allowing domain experts to distinguish different types of anomalies and common anomaly behaviors, and AI model debugging and diagnostics.
Further, to the best of the authors' knowledge, no prior research has been attempted to develop Explainable ADSs for CAN data.

The rest of the paper is organized as follows: 
Section II provides the background and related work;
Section III presents the interpretable ADS (RX-ADS);
Section IV discusses the experimental setup, results, and discussion, and finally, 
Section V concluded the paper. 

\section{Background and Related Work}

This section first discusses the Data used for training the presented RX-ADS: CAN, a widely used communication protocol for in-vehicle communication.
Then we discuss the current work on IDSs developed using CAN data. 
Finally, we discuss the background of adversarial machine learning and its applications.

    \subsection{CAN Protocol}
    Controller Area Network (CAN) is the most widely used standard bus protocol for in-vehicle communication.
    It enables efficient communication between Electronic Control Units (ECUs).
    It is a broadcast-based protocol that allows multi-master communication, and every node can initiate communication to any other node in the network. 
    Thus, CAN frames do not contain a destination address unlike other protocols.
    In CAN protocol, each ECU is able to sends messages to the vehicle communication network using data frames \cite{8514157}.
    ECUs send frames with their ID number, and the ECU on destination identifies messages by the sender ID included in the frame.
    The collision of messages and data is avoided by comparing the message ID of the node; the highest priority frame has the lowest ID. 
    CAN is proved to have many advantages, including reducing wiring cost, low weight, low complexity, and operating smoothly in an environment where electromagnetic disturbance factors exist \cite{8476919}.
    
    \begin{figure}[t]
        \centering
        \includegraphics[scale=0.55]{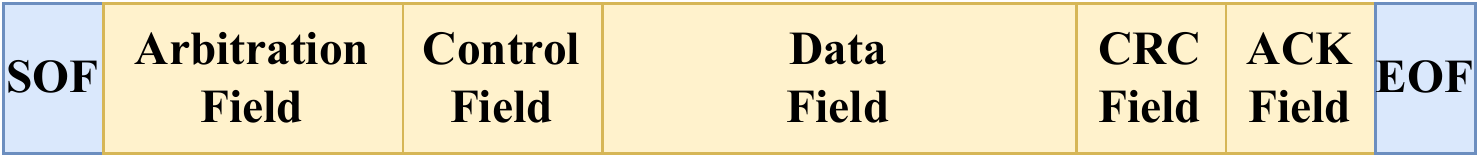}
        \caption{CAN data frame}
        \label{fig:candataframe}
    \end{figure}

    CAN protocol operates with four main types of frames: the data frame, the remote frame, the error frame, and the overload frame \cite{8476919}.
    Most of the communication happens using CAN data frame.
    The structure of the CAN data frame is presented in Figure \ref{fig:candataframe}, which consists of several common fields that are explained below \cite{8476919}.
    \begin{itemize}
        \item SOF (Start of Frame) - indicates the beginning of the frame.  
        \item Arbitration Field - is composed of message Id and RTR (Remote Transmission Request) bit. Depending on RTR state the frame will be identified as data or remote frame. During communication frames are prioritized using the ID of the frame.
        \item Control Field - sends the data size 
        \item Data Field - the actual data that node wants to send using a data frame, this field can have 0-64 bits. 
        \item CRC Field - contains 15-bit checksum that is used for error detection 
        \item Ack Field -  is used to acknowledge that a valid CAN farme was received by sending a dominant state. 
        \item EOF (End of Frame) - indicates the ending of the frame 
        
    \end{itemize}

    \begin{figure*}[h]
    \centering
    \includegraphics[scale=0.7]{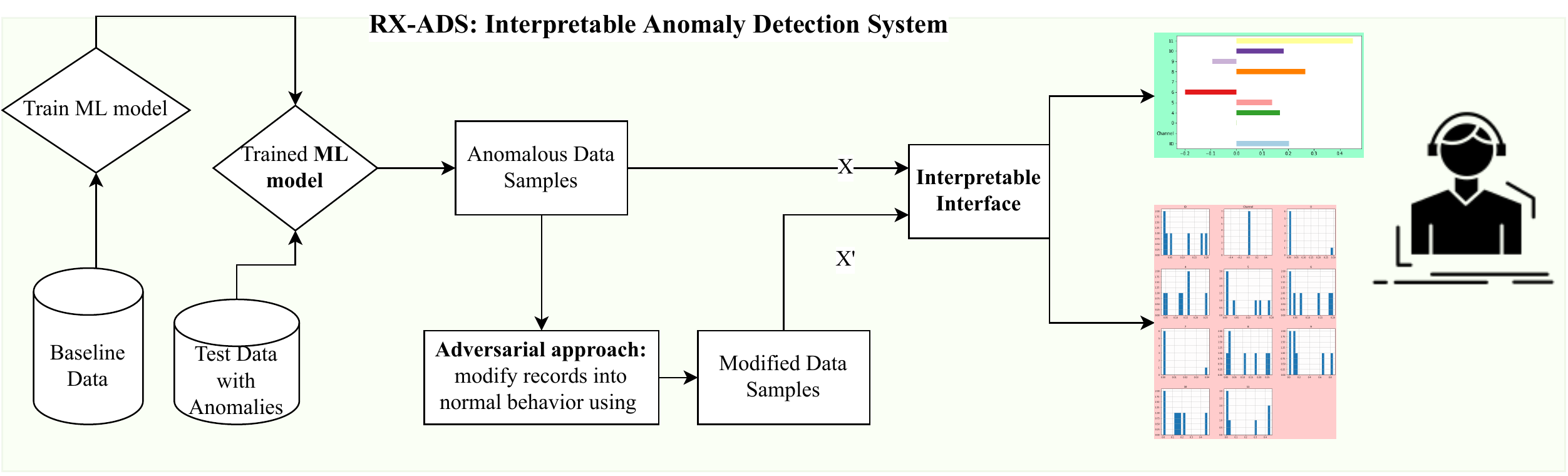}
    \caption{RX-ADS: Interpretable Anomaly Detection System Framework}
    \label{fig:figure2}
\end{figure*}
    
    \subsection{Anomaly detection using CAN data}
    
    Modern vehicles highly rely on ECU communication. Thus CAN has become the standard protocol for facilitating the data exchange between ECUs.
    While CAN protocol has many advantages, it also consists of security flows such as lack of authentication, vulnerability for various attack vectors, and lack of encryption technologies \cite{DBLP:journals/corr/abs-2012-14600}. 
    In the last couple of years, there has been a surge in research addressing security and vulnerabilities of CAN protocol, most recent work proposes different IDSs methods. 
    This subsection discusses existing IDS work on CAN data, specifically focusing on neural network methods and feature extraction techniques.
    
    To develop CAN IDSs, there are four types of feature extraction approaches has been tested in the literature \cite{DBLP:journals/corr/abs-2012-14600}.  
    First, \textit{frequency or time-based} features where timing between CAN frames and sequencing of CAN frame IDs were used for developing CAN IDSs.
    In \cite{8416254}, the broadcast time interval for each ID within a window (a discrete, non-overlapping, contiguous set of CAN frames) of can frames were calculated. Similar approach has used in \cite{10.1145/3064814.3064816}, where they calculate the signal co-occurrence time of IDs to calculate the absolute-error from expectation for identifying intrusions.
    Second feature extraction approach is \textit{Payload-based} approach where message content bits are directly used for building CAN IDSs \cite{DBLP:journals/corr/abs-2012-14600}. 
    Third approach is \textit{Signal based} approach where message content is decoded into signal before feeding into the IDS. For example in \cite{article_CANNET}, payload bits are encoded before feeding into Neural network architecture for detection intrusions in CAN.
    Finally, the forth approach is \textit{Physical side channels} where physical attributes such as voltage, temperature are used to detect intrusions \cite{DBLP:journals/corr/abs-2012-14600, 8306904}.
    Other than these four approaches, some IDSs have developed using rule based methods where characteristic of CAN communication was encoded into rules for detecting intrusions \cite{DBLP:journals/corr/abs-2012-14600}. The main drawback of rule based methods is that they require domain knowledge and manual rule generation, which are challenging and time-consuming tasks.

    Neural Network (NN) based CAN IDSs are mainly developed by encoding the characteristics of CAN communication into a set of features and training NN algorithms on these extracted sets of features. 
    The features are extracted to ensure capturing the normal behavior patterns of CAN bus communication apart from abnormalities. 
    For example, in \cite{article_CANNET}, Long Short Term Memory (LSTM)  and AE-based unsupervised IDS was developed for detecting intrusions. This architecture consists of a neural network architecture where CAN data from each ID type is presented to its assigned LSTM. The results of LSTM networks are aggregated into AE NN. They have tested their approach on Synthetic CAN data only. 
    A similar approach was used in \cite{7796898} where they used LSTM for detecting anomalies. However, they have not aggregated results of multiple LSTMs using AE. 
    In \cite{Kang2016IntrusionDS}, deep NN-based IDS was presented, which used deep belief network (DBN) pre-training methods for initial parameter optimization. This approach is a supervised approach where labeled data was required to build the IDS.
    Convolutional Neural Network (CNN) based supervised CAN IDS was proposed in \cite{9322395} where they have tested their system on a real CAN data set. They have directly fed information in CAN frames as features for the training of CNN.

    \subsection{Adversarial Machine Learning}

    Adversarial samples are generally referred to as malicious input samples designed to fool machine learning algorithms \cite{DBLP:journals/corr/KurakinGB16a}. 
    These samples are typically created by adding a slight modification into real data samples, such that the outcome of a machine learning model for crafted samples will be different than the real sample \cite{8591457}.
    Typically, machine learning models are vulnerable to these generated adversarial samples, resulting in unintended or incorrect outcomes.
    Generally, adversarial samples are generated to maximize the impact on the model while minimizing the ability to identify the adversarial sample apart from a real sample.  This ensures by keeping the adversarial sample inside the domain of valid inputs.

    Adversarial machine learning has been widely used for exposing vulnerabilities in critical infrastructures \cite{8591457}. 
    The positive or negative impact of adversarial ML depends on the purpose of the use of these samples.
    For example, an attacker can use these samples can be used to gain information on a trained ML model, information on the data set the model trained on, and attack a model. 
    This results in possible privacy invasion, safety failures, data corruptions, and model theft \cite{8591457, 9283867}.
    On the other hand, adversarial machine learning also can be used for improving the performance of machine learning models.
    For example, it can be used to eliminate undefined behaviors of ML models, exploit vulnerabilities, assess model robustness and improve generalization  \cite{8591457, 9283867, inproceedings111, BIGGIO2018317}. 
    This paper uses adversarial ML to interpret CAN ADS, which helps with interpreting the decision making process of black-box NN models and help with NN model debugging and diagnostics.


\section{RX-ADS Methodology}

This section discusses the development of RX-ADS. 
First, we extract a set of cyber features from raw CAN communication data, i.e., converting the raw CAN data into a data format that can feed into ML algorithms. 
Then, it trains an ML model using baseline data, i.e., data that represent the normal behavior of the system. 
Then the trained ML model was tested using various abnormal scenarios.
Once it identifies abnormal samples, these samples are modified using an adversarial approach, i.e., it performs the minimum modification required to change the anomalous records ($ x'$) into normal/baseline records ($ x $). 
The difference between $ x$ and $x'$ is used to explain the ADS outputs, illustrating the most relevant features that lead to anomalous behaviors. 
The proposed approach is illustrated in Figure \ref{fig:figure2}.
The individual system components are explained below.

\begin{table*}[h]
	\caption{\label{table:window_feature_extraction} Presented feature extraction method}
	\centering
	\resizebox{15cm}{!}{
    \begin{tabular}{r l} \hline
        \multicolumn{2}{l}{Algorithm I: Extract features}\\ \hline
        \multicolumn{2}{l}{Inputs: Dataset $(X)$, Time window size $(winSize)$, Possible set of signal IDs $(IDList)$ }\\ 
        \multicolumn{2}{l}{Outputs: Window features }\\ \hline
1:          & $ startTime = 0$ \\ 
2:          & $ listRecords = [] \leftarrow $ = Initialize a list to store window features \\
3:          & $ endTime = Timestamp of last record of X \leftarrow $ = Store the last timestamp of the dataset \\
4:          & \% Calculating features for each overlapping time window  \\ 
5:          &  \textbf{while} $startTime < endTime $ do  \\ 
6:          & \hspace{5mm} $ windowMessages  \leftarrow $ Extract massages from $X$ where timestamp is within range $ startTime - (startTime + winSize)$ \\
7:          & \hspace{5mm} $ no\_of\_records  \leftarrow $ Number of messages in $ windowMessages$ \\
8:          & \hspace{5mm} $ no\_of\_ids  \leftarrow $ Number of unique IDs in $ windowMessages$ \\
9:          & \hspace{5mm} $ no\_of\_dlc  \leftarrow $ Number of unique data length of messages in $ windowMessages$ \\
10:          & \hspace{5mm} $ time\_interval  \leftarrow $ Minimum/Maximum/Mean timestamp differences of messages in $ windowMessages$ \\
11:          & \hspace{5mm} $ no\_of\_req\_msgs  \leftarrow $ Number of remote frames in $ windowMessages$ \\
12:          & \hspace{5mm} $ no\_of\_res  \leftarrow $ Number of response frames in $ windowMessages$ \\
13:          & \hspace{5mm} $ no\_of\_lost  \leftarrow $ Number of lost response frames in $ windowMessages$ \\
14:          & \hspace{5mm} $ ratio  \leftarrow $ Number of messages between requests and responses in $ windowMessages$ (Minimum, Maximum, Mean) \\
15:          & \hspace{5mm} $ reply\_time\_interval  \leftarrow $ Minimum/Maximum/Mean timestamp differences of requests and responses in $ windowMessages$  \\
16:          & \hspace{5mm} $ 0000  \leftarrow $ Number of high priority messages (ID=0000) in $ windowMessages$  \\
17:          & \hspace{5mm} $ no\_XXXX  \leftarrow $ Number of messages with $ID=XXXX$ in $ windowMessages$  \\
18:          & \hspace{5mm} $ payload\_pX\_XXXX  \leftarrow $ Mean signal values of payload signal $x$ from messages with $ID=XXXX$ in $ windowMessages$  \\

19:          & \hspace{5mm} $ startTime += (winSize/2) \leftarrow $ Calculate start time of next window \\
20:	        & \textbf{end while} \\
\hline 
\hline 
\end{tabular}
}
\end{table*}

\begin{table*}[h]
\centering
\caption{Feature List \label{table:feature_list}}
\begin{tabular}{l|l}
\hline
\hline
\textit{\textbf{Feature}}                       & \textbf{Description} \\ \hline\hline
\textit{no\_of\_records}                        & Number of CAN messages                                                 \\ \hline
\textit{no\_of\_ids}                            & Number of unique CAN message IDs                                       \\ \hline
\textit{no\_of\_dlc}                            & Number of unique CAN message payload lengths                           \\ \hline
\textit{time\_interval}                    & Time interval between messages (minimum, maximum and mean)             \\ \hline
\textit{no\_of\_req\_msgs}                      & Number of request frames                                               \\ \hline
\textit{no\_of\_res}                            & Number of responses                                                    \\ \hline
\textit{no\_of\_lost}                           & Number of lost responses                                               \\ \hline
\textit{ratio (min, max, mean)}                 & Number of messages between request frame and response frame            \\ \hline
\textit{instant\_reply\_count}                  & Number of instant reply messages                                       \\ \hline
\textit{reply\_time\_interval (min, max, mean)} & Time difference between request frame and corresponding response frame \\ \hline
\textit{high\_priority\_count/ 0000}            & Number of high priority messages                                       \\ \hline
\textit{no\_XXXX}                               & Number of messages with ID XXXX                                        \\ \hline
\textit{payload\_P1\_XXXX}                      & Mean payload of signal P1 with ID XXXX                                 \\ \hline\hline
\end{tabular}

\end{table*}

\subsection{Feature Engineering}
This subsection describes the window-based feature engineering approach where we generate a set of cyber features from raw CAN frames.
The extracted features are used to feed the ML algorithm for training.
This is motivated by widely used window-based network flow feature extraction methods in industrial control ADSs \cite{8473535}.
The main goal of this approach is to extract a set of features using a set of CAN messages within a defined-sized time window.
These features are selected based on the available literature on CAN bus data \cite{9463874,8514157}. 
The sliding window-based feature extraction algorithm is presented in Algorithm I in Table \ref{table:window_feature_extraction}. 
For each dataset, we used a time window and extracted features using the CAN data frames within that window. 
Overlaps between two windows are kept as half of the window size.
In this experiment, window features are extracted using different time window sizes ($winSize$).

The set of extracted features with their description is presented in Table \ref{table:feature_list}. 
These features are extracted to represent the fluctuation in normal behavior in the CAN communication data.
All or some of the features are extracted for each tested dataset.
Then, the extracted features were fed into the AE model for building data-driven ADS.

\subsection{Data-Driven Machine Learning Model}
As discussed in the introduction, Autoencoder (AE) NN model was used to learn the normal/baseline behavior of the system.
Specifically, we used deep ResNet AE (RAE) architecture to avoid possible performance degradation, and easy parameter optimization \cite{9373350}.  
AE has an encoder and decoder, each consisting of multiple hidden layers. 
Training of the model consists of two stages, the encoding stage, and the decoding stage. The encoding stage transforms the input data into an embedded representation, whereas in the decoding stage, the embedded representation is reproduced back to the original input record (reconstruction). 
Encoding and decoding functions are non-linear transformation functions; in this experiment, we used a Sigmoid function. 
The loss function ($J_{\theta }$) of the AE model is computed using the difference between the input ($x$) and the reconstruction ($x'$). Thus reconstruction error of the AE is calculated as follows, 
\begin{equation}
    J_{\theta } = \frac{1}{T} \sum_{i=1}^{T} \| x_{i} - x'_{i} \| ^{2}
\end{equation}
where $ x_{i} $ is the  $ i $ th input sample, $ x'_{i} $ is the reconstruction for $ i $th input sample, $ \theta $ denotes the set of parameters of the AE (weights and biases). 

During training, AE is trained with data coming from the normal behavior of the system. Therefore, it only learns the possible normal behaviors of the system.
When unseen records are presented to the trained AE, the amount of reconstruction error indicates how much the presented data differs from the learned normal behavior.
A threshold value is defined to identify possible anomalies. The data records were detected as anomalies if the reconstruction error was higher than the defined threshold value. Thus, given data record $x_i$ is detected as anomaly ( $ y = 1$) or normal ($y=0$) as follows, 
\begin{equation}
    J_{\theta,i} =  \| x_{i} - x'_{i} \| ^{2}
\end{equation}
\begin{equation}
    y = \left\{ \begin{array}{cl}
1 & : \ J_{\theta,i}  \geq th \\
0 & : \ J_{\theta,i}  < th
\end{array} \right.
\end{equation}
where $ J_{\theta,i}$ is the reconstruction error of $i$th data record, $th$ denotes the threshold value, and $y$ represents predicted label: anomaly or not. The threshold value is optimized based on the training baseline data, i.e., the threshold value should capture the baseline data boundary, capturing the normal behavior fluctuations. 

We experimented with different time window sizes and different RAE architecture. 
L1 regularized RAE architecture was used as some of the features can result in data sparsity.
Mean squared error was used as the loss function. 
A different number of hidden layers sizes were tested, and the paper presents the best results.
We used the proposed feature engineering method to extract a set of cyber features for all datasets.
We divided the baseline/normal data into two sets (train/test) with a 0.7/0.3 ratio.
The data was scaled to the 0-1 range. 
Once the RAE model is trained with baseline data, the reconstruction errors on train data were used to define an error threshold by keeping 99.9999\% train data within the defined threshold. 
The trained RAE's (ADS) performance was evaluated using the test baseline data and abnormal/attack data.

\subsection{Modifying Anomalous Samples: Adversarial Approach}
This paper uses adversarial Machine Learning (ML) to understand why a given sample is detected as an anomaly. 
Further, these understandings of data samples are aggregated to understand different scenarios on how they are different from each other and what feature changes are prominent in each scenario.  
Thus, this adversarial ML approach aims to understand the decision boundary of normal data and to understand how abnormal scenarios affect the system. 
This is important for assurance and validation of AI systems.

The concept of adversarial sample generation was used to find the minimum modification needed to change the anomalous sample into a normal behavior sample.
This is achieved by finding an adversarial sample $x''$ that is detected as a normal sample with the given $th$ while minimizing the distance between real sample $x$ and adversarial/modified sample $x''$.  
\begin{equation}
   \underset{x''}{min}
 \left\| x' - x'' \right\|^{2}
\end{equation}

\begin{equation}
\begin{array}{rcl}
 s.t: J_{\theta,x''} \le  th \\
x_{min} <= x'' <= x_{max} \\
\end{array}
\end{equation}

We constrain the adversarial sample $x''$ o be inside the bounds ($x_{min}, x_{max}$). These bounds are defined using the training data, ensuring that the adversarial samples are inside the domain of data distribution.

\subsection{Interpretable Interface}

The presented interpretable interface generates explanations for detected anomalies. 
For generating explanations, RX-ADS uses the identified anomalous samples as references, then uses Eq 4 and 5 to find the adversarial samples with minimum modifications. 
Explanation generated under two categories:
\begin{itemize}
    \item Explanations for individual Anomaly Samples: These explanations are generated by calculating the difference between the anomaly sample and the closest adversarial sample ($ x - x'' $) and visualizing it using a bar chart. This bar chart shows the deviation of the anomaly sample from what the model learned as normal behavior. Domain experts can analyze these graphs quantitatively to understand the root courses of a given anomaly.  For example, identifying cyber anomalous behavior which leads to a physical impact in the vehicle. 
    \item Explanations for global anomalous behavior: Explanations generated for anomalous data records are aggregated to understand common anomaly behaviors in the system. It allows domain experts to distinguish different types of anomalies and common anomaly behaviors. 
\end{itemize}

tying a cyber root cause to a physical impact.


\section{Experiment, Results, and Discussion}
The proposed system was tested against two popular CAN protocol benchmark datasets. 
Both datasets contain CAN bus data representing normal behavior and abnormal/attack behaviors. 
This section discusses RX-ADS results and discussion for each dataset.

\begin{table}[]
\caption{\label{table:ads_performance_comparison} OTIDS dataset: RX-ADS anomaly detection comparison with recent literature}
\centering
\begin{tabular}{l|l|l|l}
\hline
\hline
\textbf{Approach}                  & \textbf{Normal Behavior} & \textbf{DoS}   & \textbf{Fuzzy} \\ \hline \hline
\textit{\textbf{HIDS \cite{9463874} }} & 100\%                    & 100\%          & 100\%          \\ \hline
\textit{\textbf{OCSVM}}           & 99.77\%           & 100\% & 100\% \\ \hline
\textit{\textbf{LOF}}           & 99.32\%           & 100\% & 100\% \\ \hline
\textit{\textbf{RX-ADS}}           & \textbf{100\%}           & \textbf{100\%} & \textbf{100\%} \\ \hline \hline

\end{tabular}
\end{table}

\begin{figure*}[h]
  \centering
  \subfigure[DoS]{\includegraphics[scale=0.23]{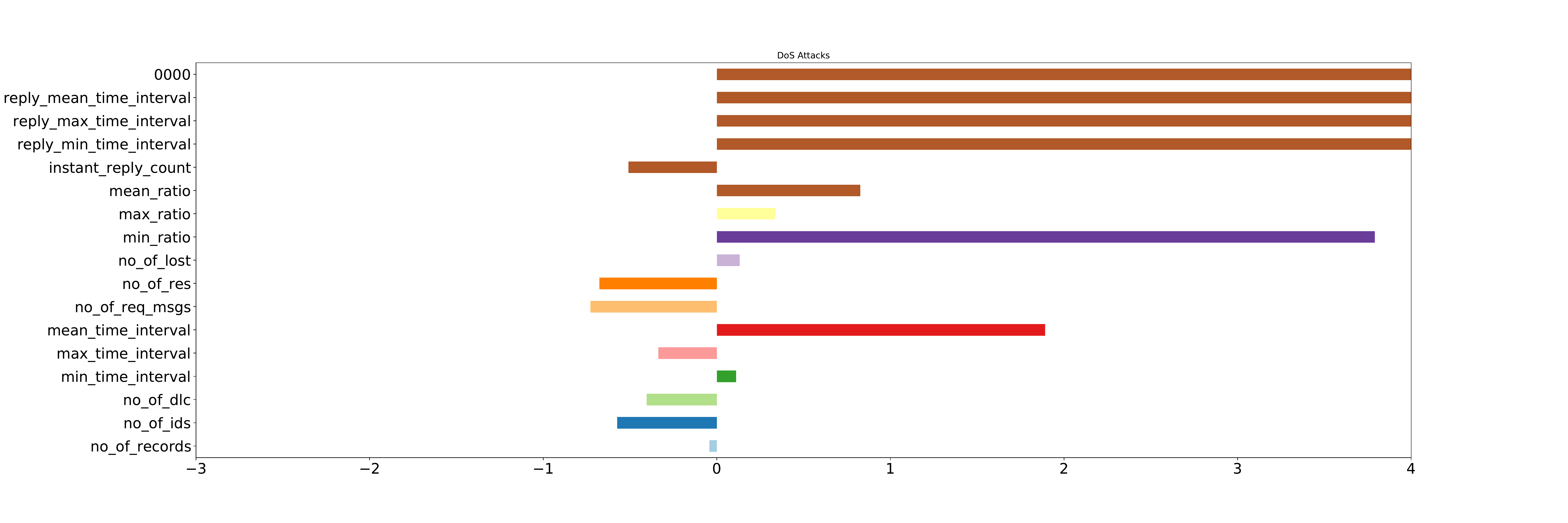}}\quad
\\
  \subfigure[Fuzzy]{\includegraphics[scale=0.23]{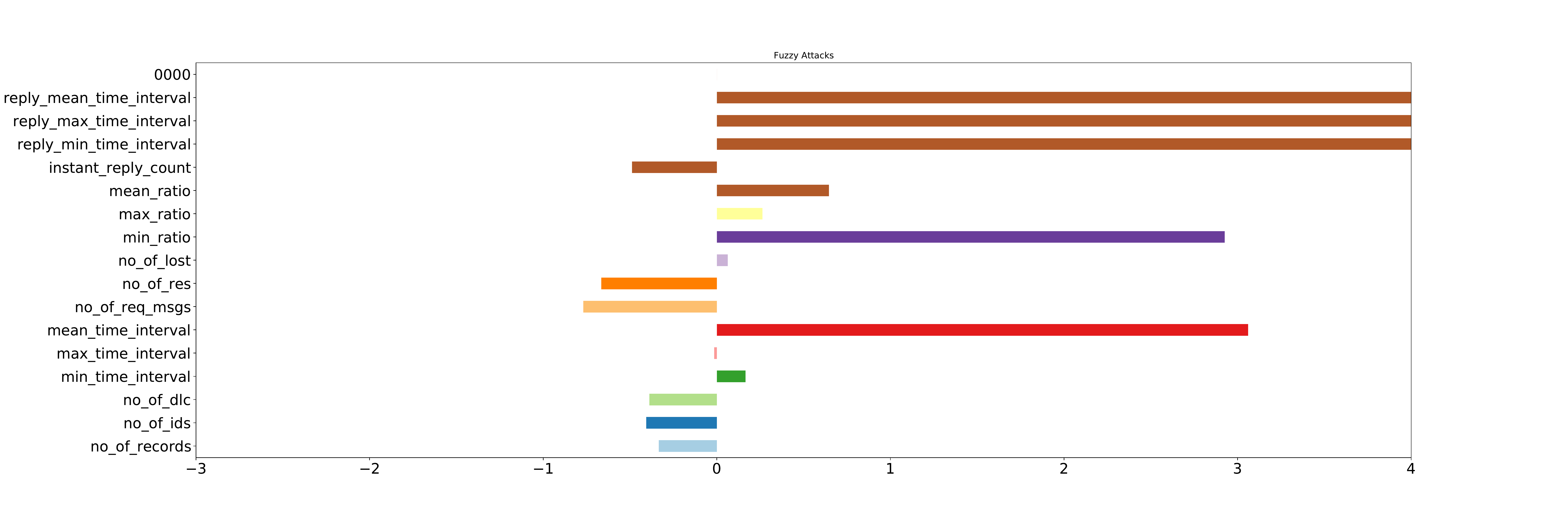}}\quad
\\
   
  \caption{OTIDS dataset: Outcomes generated using adversarial approach for DoS records and Fuzzy records \label{fig:global_explanation} }
\end{figure*}

\subsection{RX-ADS vs. HIDS (OTIDS dataset)}
This section compares the anomaly detection performance between RX-ADS and HIDS algorithms. The Histogram-based IDS (HIDS) is the state-of-the-art approach for detecting intrusions in the OTIDS dataset.

The OTIDS benchmark CAN dataset was released by Hacking and Countermeasures Research Lab (HCRL) \cite{8476919}. 
It contains real CAN data collected from a Kia Soul vehicle in normal behavior and during a set of attacks: DoS, Fuzzy, and impersonate.
With the release of this dataset, the authors also proposed an Offset Ratio and Time interval-based Intrusion Detection System (OTIDS).
Their approach uses offset ratio and time interval of remote frame responses to identify intrusions in CAN communication. 

This experiment used baseline, DoS, and Fuzzy data to simplify the experiment and compare the explanations. 
DoS Attacks have been implemented by injecting messages of ‘0x000’ CAN ID in a short cycle.
Fuzzy Attacks have been implemented by injecting messages of spoofed random CAN ID and DATA values.
All the features except $payload\_PX\_XXXX$ were extracted for this dataset. 
This was performed due to the available domain knowledge on this dataset shows that it is possible to identify abnormalities by using only remote request and response-based features.
It has to be noticed that this is the only open dataset with remote frames and responses, such that it is important to experiment on how this information is essential for anomaly identification.

\begin{table*}[h]
\caption{\label{table:otids_attack_vs_normal} OTIDS dataset: Natural interpretation of abnormal communication compared to normal}
\centering
\begin{tabular}{l|l|l}
\hline \hline
\textbf{Characteristics of communication}                                                                                                                     & \textbf{Normal communication} & \textbf{Abnormal communication (DoS and Fuzzy)} \\
\hline \hline
\textit{High priority CAN frames with ID 0000}                                                                                                                & Low                           & High                                            \\
\textit{Min/max/mean time interval between remote and response messages}                                                                                      & Low                           & High                                            \\
\textit{\begin{tabular}[c]{@{}l@{}}Number of CAN messages between the remote frame and\\  its corresponding response frame (min/max/mean ratio)\end{tabular}} & Low                           & High                                            \\
\textit{Higher number of lost response messages}                                                                                                              & Low                           & High                                            \\
\textit{Number of instant reply, request, and response  messages}                                                                                             & High                          & Low                                             \\
\textit{Number of unique IDs, number of records, and unique DLC values}                                                                                       & High                          & Low                                             \\
\hline \hline
\end{tabular}
\end{table*}

\subsubsection{Anomaly Detection System Performance}

We experimented with different millisecond time window sizes :0.01, 0.02, 0.05, 0.1, 0.2, 0.5, 1, 2, 5.
We found that when the time window is too small ($<$0.02), the performance of the fuzzy attack detection rate decreases. 
Further, the baseline accuracy was reduced if the time window is too large (0.1$>$). 
The best-observed result was observed for 0.05 milliseconds window size, which is recorded in this section.
Table \ref{table:ads_performance_comparison} shows the detection performance of presented RX-ADS compared to recent state-of-the-art IDSs on OTIDS dataset: Histogram-based approach (HIDS) presented in \cite{9463874}. 
We also implement two widely used anomaly detection algorithms: Local Outlier Factor (LOF) and One-Class SVM (OCSVM).

It can be seen that RX-ADS shows comparable performance with the state-of-the-art approach on this dataset. Both HIDS and RX-ADS used window-based feature exaction methods.
The HIDS uses a fixed number of CAN frames as a window, whereas RX-ADS uses CAN messages within a fixed time window.
It has to be noted that the HIDS uses the K-Nearest Neighbor (KNN) algorithm for performing multi-class classification. 
Thus it requires labeled data from all the classes (normal, DoS, and Fuzzy) for training. 
However, RX-ADS only requires data from normal behavior for training, which is advantageous as the data labeling is expensive \cite{9536751,9373350}.
Further, to the best of our knowledge, none of the IDSs proposed on this dataset use an explainable approach.
It has to be noted that the goal of this approach is not only to detect anomalies but also to interpret the reasons behind anomalies (interpret the decision-making process of black-box AI models).


\subsubsection{Explanation generation}

Figure \ref{fig:global_explanation} shows the deviations calculated using the adversarial method for two types of abnormal/attack behaviors (DoS and Fuzzy).
As we discussed before, these deviations of feature values not only explain the behavior of attacks compared to normal behavior but also help with distinguishing different types of attacks. 
To make the comparison easy, deviations for two types of attacks were presented with the same scale.
Explanations for attack behaviors can be naturally interpreted as compared to normal behavior in following tabular format in Table \ref{table:otids_attack_vs_normal}.


Related literature on this dataset confirms that the normal communication has a very low lost reply rate, a higher number of instant reply rates, and a very low/zero amount of high priority messages.
Further, a higher number of messages during normal communication results in a higher number of unique IDs, number of records (messages), and unique DLC values. 
However, during attacks, it generates high-priority messages or spoofs random messages, resulting in collisions between CAN frames.
It leads to a delay in CAN communication.
Thus, fewer records are expected during abnormal communication compared to baseline. Further, this results in a fewer number of unique DLCs, IDs, and a number of records within a window.
Thus, the identified features of attacks match the domain expert's knowledge of this dataset.

The explanation generated for two types of attacks can be compared against each other to identify distinguishing characteristics between them.
The explanations for distinguishing two behaviors can be naturally interpreted in the following manner.

DoS and Fuzzy attacks affects the system differently based on the following observations: 
\begin{itemize}
    \item DoS result in a higher number of high priority messages (0000), whereas Fuzzy does not result in high priority messages with ID 0000
    \item Min/Max/Mean ratio is higher for DoS due to high priority message communication. 
    \item No of lost response frame rate is higher for DoS.

\end{itemize}

\begin{figure*}[t]
  \centering
  \subfigure[no\_of\_lost]{\includegraphics[scale=0.16]{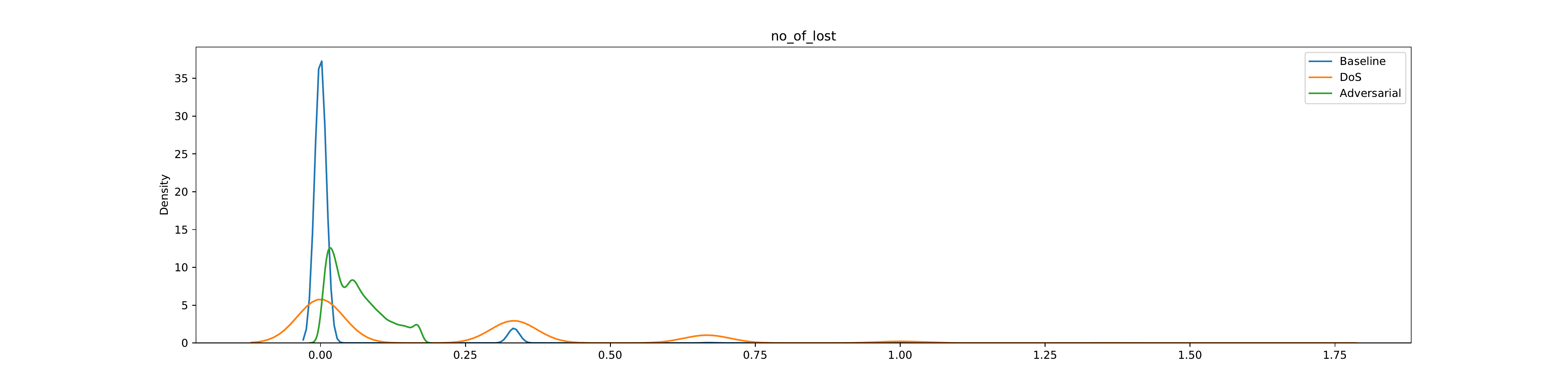}}
  \subfigure[no\_of\_lost]{\includegraphics[scale=0.16]{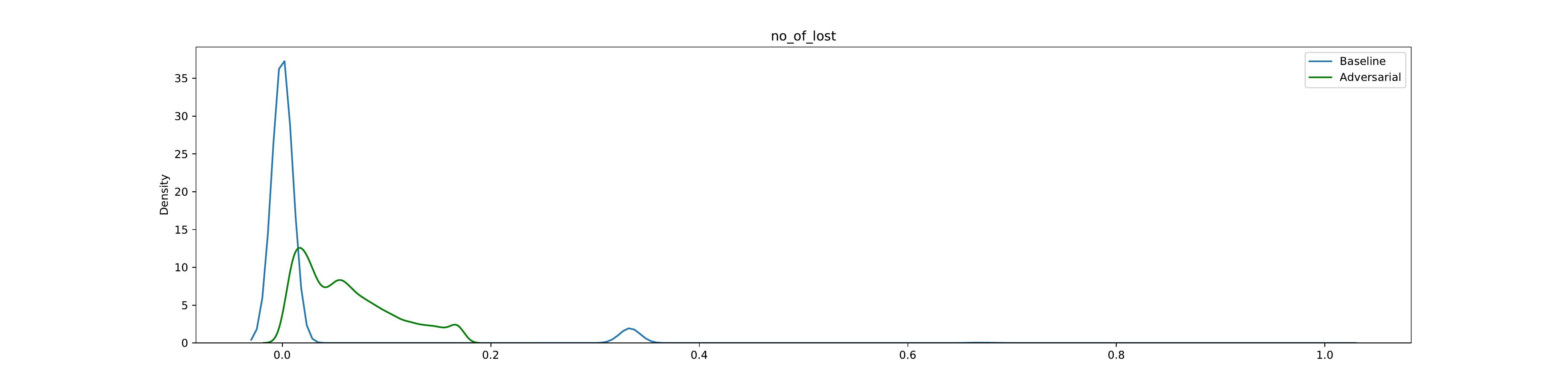}}
\\
  \subfigure[mean\_ratio]{\includegraphics[scale=0.16]{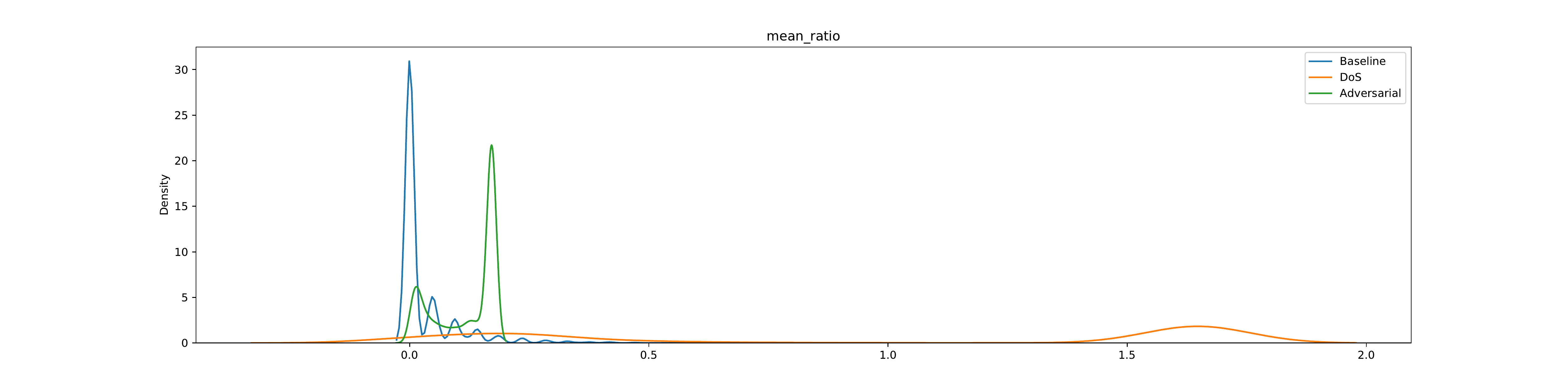}}
  \subfigure[mean\_ratio]{\includegraphics[scale=0.16]{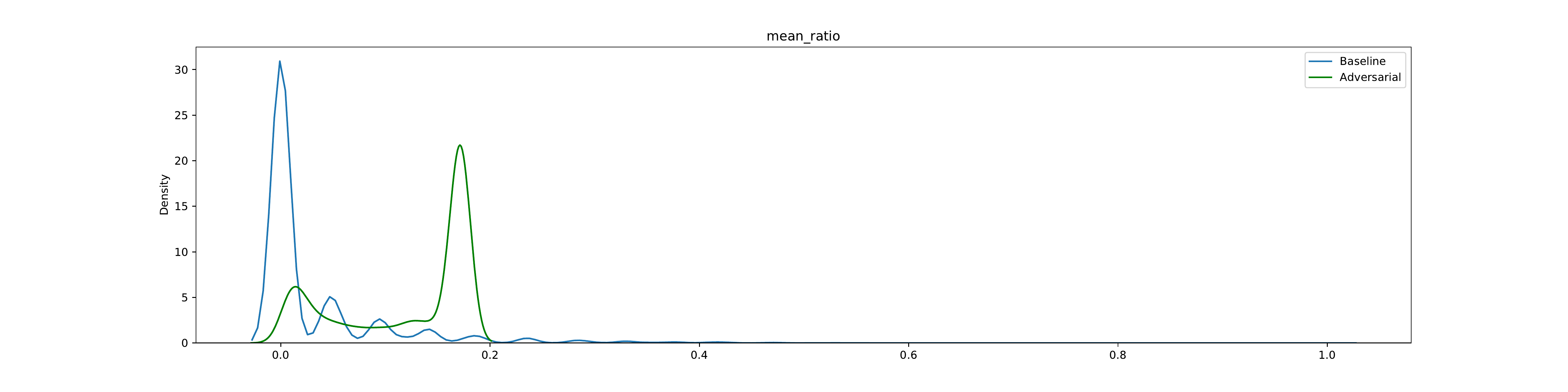}}
\\
 \subfigure[instant\_reply\_count]{\includegraphics[scale=0.16]{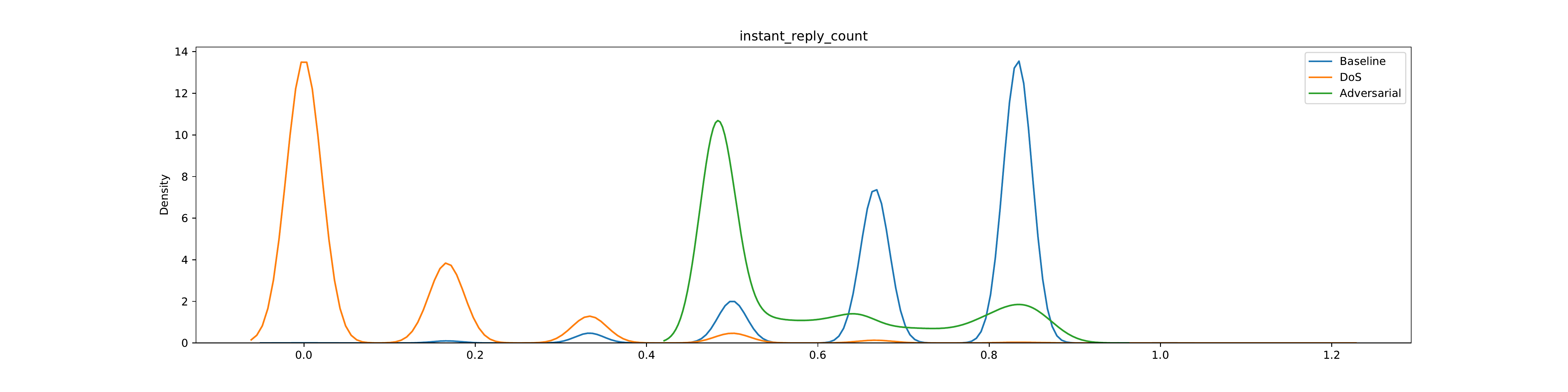}}
  \subfigure[instant\_reply\_count]{\includegraphics[scale=0.16]{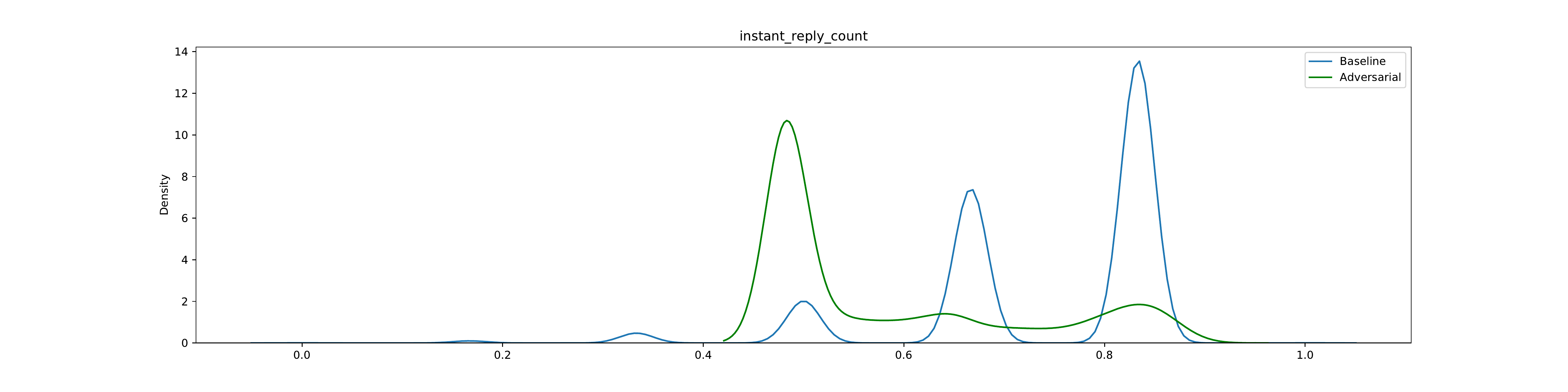}}
\\
   
  \caption{DoS: Feature value distribution of DoS data and Adversarial data compared to normal feature value distribution \label{fig:dos_feature_val_distribution} }
\end{figure*}
\begin{figure*}[t]
  \centering
  \subfigure[no\_of\_lost]{\includegraphics[scale=0.16]{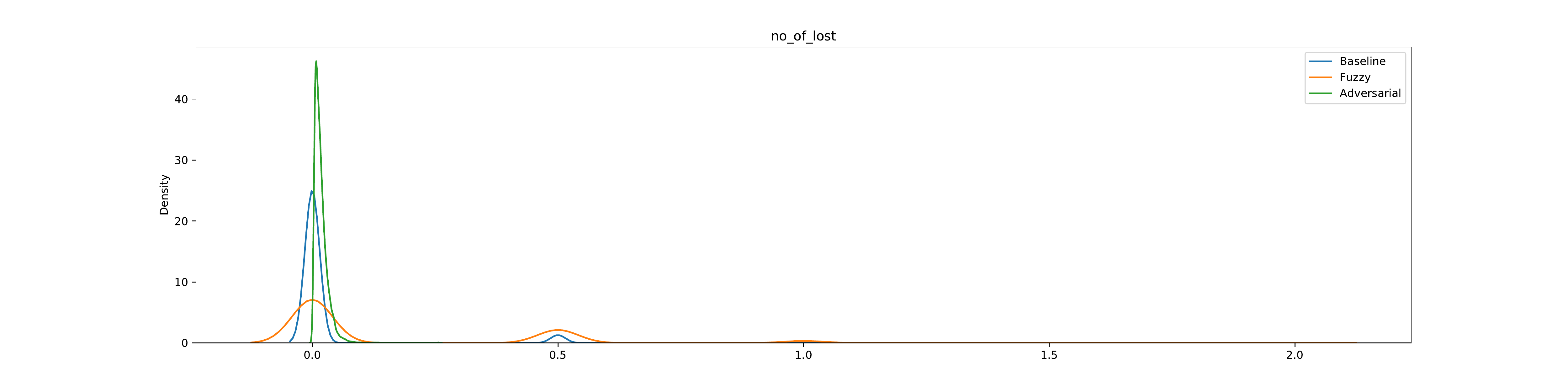}}
  \subfigure[no\_of\_lost]{\includegraphics[scale=0.16]{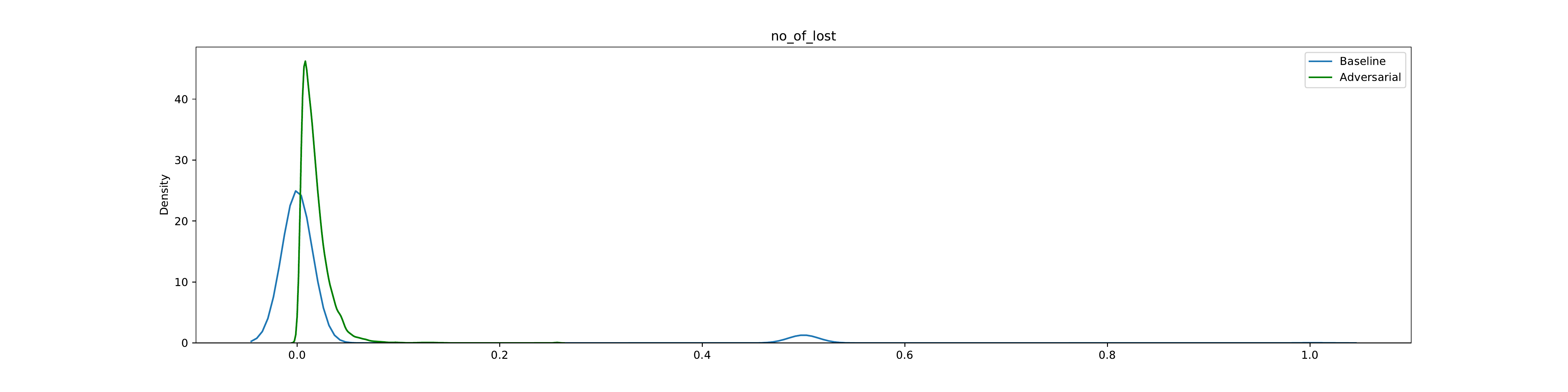}}
\\
  \subfigure[mean\_ratio]{\includegraphics[scale=0.16]{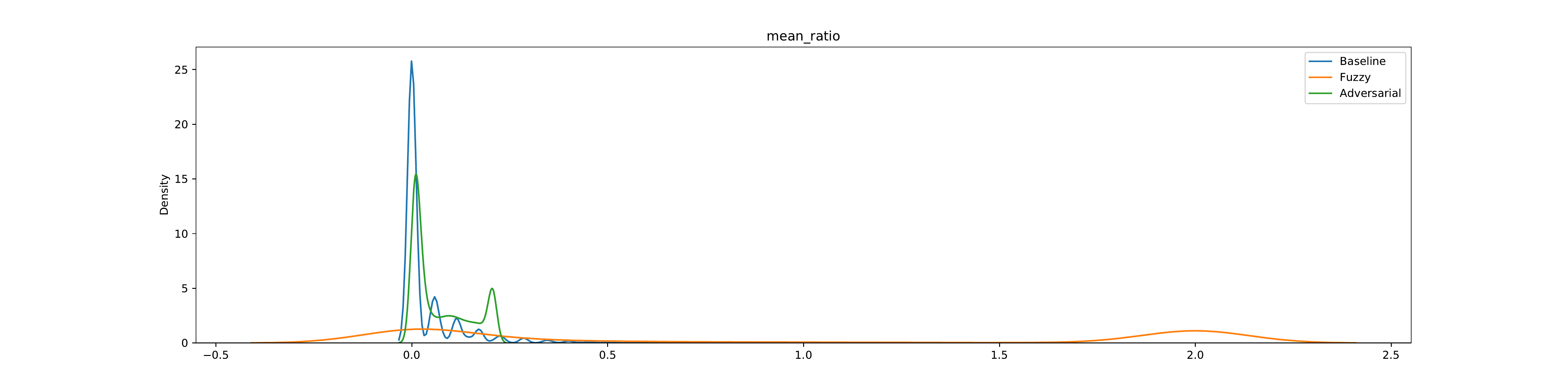}}
  \subfigure[mean\_ratio]{\includegraphics[scale=0.16]{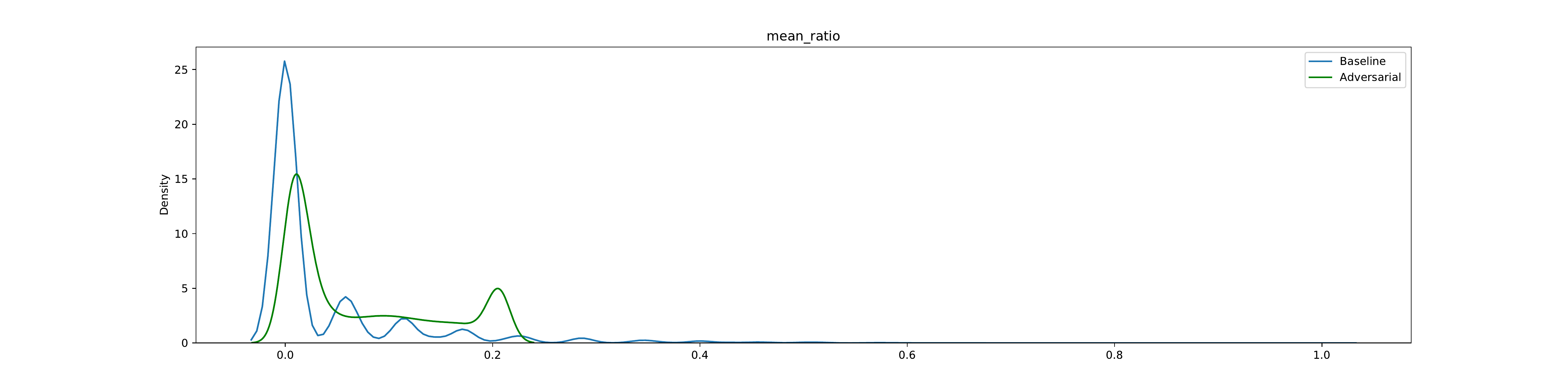}}
\\
 \subfigure[instant\_reply\_count]{\includegraphics[scale=0.16]{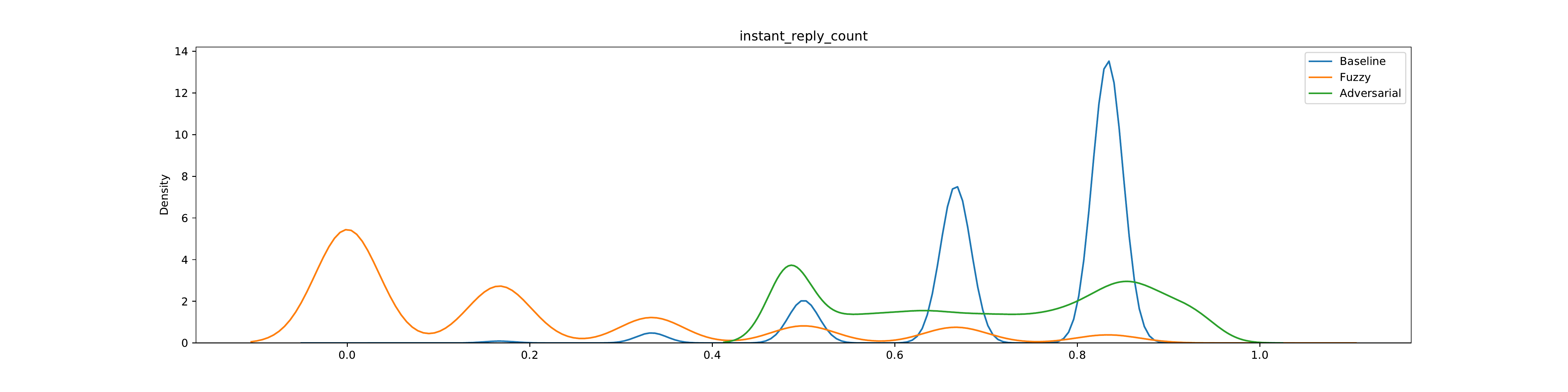}}
  \subfigure[instant\_reply\_count]{\includegraphics[scale=0.16]{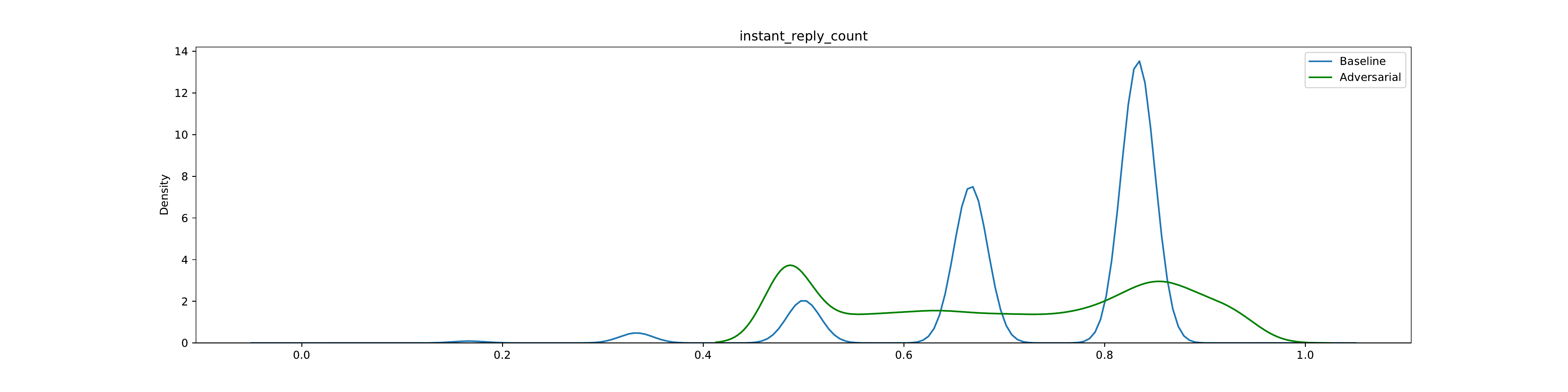}}
\\
   
  \caption{Fuzzy: Feature value distribution of Fuzzy data and Adversarial data compared to normal feature value distribution \label{fig:fuzzy_feature_val_distribution} }
\end{figure*}

Once adversarial samples are generated, feature value distribution of baseline, attacks, and adversarial records also give insights into how different features behave under abnormalities. 
Figure \ref{fig:dos_feature_val_distribution} illustrates the feature behavior for selected features under DoS attacks.
It can be seen that many of the identified features deviate from the baseline behavior with different magnitudes (Orange line). 
However, generated adversarial samples (green line) have a much closer feature value distribution than the baseline (blue line).
Feature value distribution during Fuzzy attacks is also presented in Figure \ref{fig:fuzzy_feature_val_distribution}, which also shows similar behavior. 

\subsection{RX-ADS vs. GIDS and HIDS (Car Hacking dataset)}
This section compares the anomaly detection performance between RX-ADS with GIDS and HIDS algorithms. 
The Generative neural network-based IDS (GIDS)  is the state-of-the-art approach for detecting intrusions in the Car Hacking dataset.

The Car Hacking dataset is the most recent dataset released by the Hacking and Countermeasures Research Lab (HCRL). This dataset contains real CAN data collected from Hyundai YF Sonata. 
It contains a baseline data file, a data file with DoS attacks, a data file with Fuzzy attacks, and a data file with Spoofing attacks. 
CAN frame structure is very similar compared to their first released OTIDS dataset. 
Remote frames indicators are only included in baseline data; thus, this experiment ignores remote information bit from CAN frames. 

In this experiment, we used only baseline, DoS and Fuzzy CAN data. 
All the features except payload PX\_XXXX
were extracted for this dataset. This was performed due to the available domain knowledge confirming that it is possible to identify intrusions only using timing information and ID frequencies. 
This dataset seems to be the most widely used dataset in the CAN IDS literature \cite{DBLP:journals/corr/abs-2012-14600}. 
Initial data analysis indicated large gaps between CAN frames during attacks \cite{Berger2018ComparativeSO,DBLP:journals/corr/abs-2012-14600}. Our analysis also confirmed this behavior. 
Hence we trimmed attack datasets before using them for the experimentation.

    \subsubsection{Anomaly Detection System Performance}
    
\begin{table}[]
\caption{\label{table:ads_performance_comparison_carHacking} Car Hacking Dataset: RX-ADS anomaly detection comparison with recent literature}
\centering
\begin{tabular}{l|l|l|l|l|l}
\hline \hline
\textit{\textbf{Method}}                     & \textbf{Data}                                                     & \textbf{Accuracy} & \textbf{Precision} & \textbf{Recall} & \textbf{F1} \\ \hline \hline
\multirow{2}{*}{\textit{\textbf{HIDS}}} & \textit{DoS}                                                      & 97.28             & 100                & 96.2            & 98.06       \\ \cline{2-6} 
                                             & \textit{Fuzzy}                                                    & 95.17             & 99.55              & 94.3            & 97.18       \\ \hline
\multirow{2}{*}{\textit{\textbf{GIDS}} \cite{8514157}}      & \textit{DoS}                                                      & 97.9              & 96.8               & 99.6            & 95.42           \\ \cline{2-6} 
                                             & \textit{Fuzzy}                                                    & 98.0              & 97.3               & 99.5            & 98.39           \\ \hline
\multirow{3}{*}{\textbf{RX-ADS}}             & \textit{\begin{tabular}[c]{@{}l@{}}Baseline \\ Test\end{tabular}} & 100               & -                  & -               & -           \\ \cline{2-6} 
                                             & \textit{DoS}                                                      & 99.47             & 99.6               & 99.74           & \textbf{99.67}       \\ \cline{2-6} 
                                             & \textit{Fuzzy}                                                    & 99.19             & 99.39              & 99.63           & \textbf{99.51}       \\ \hline \hline
\end{tabular}
\end{table}

\begin{table*}[h]
\centering
\caption{\label{table:carhacks_attack_vs_normal} Car Hacking dataset: Natural interpretation of abnormal communication compared to normal}
\begin{tabular}{l|l|l|l}
\hline \hline
\multirow{2}{*}{\textbf{Characteristics of communication}}               & \multirow{2}{*}{\textbf{Normal communication}} & \multicolumn{2}{l}{\textbf{Abnormal communication}} \\
                                                                         &                                                & \textbf{DoS}            & \textbf{Fuzzy}            \\
\hline \hline
\textit{High priority CAN frames with ID 0000}                           & Low                                            & High                    & Low                       \\
Odd ID can frames within a window                                        & Low                                            & Highest                 & High                      \\
\textit{Min/max/mean time interval between remote and response messages} & Low                                            & High                    & Highest                   \\
\textit{Number of unique CAN frames within a window}                     & Low                                            & Low                     & High                     \\
\hline \hline
\end{tabular}
\end{table*}

\begin{figure*}[ht!]
  \centering
  \subfigure[DoS]{\includegraphics[scale=0.23]{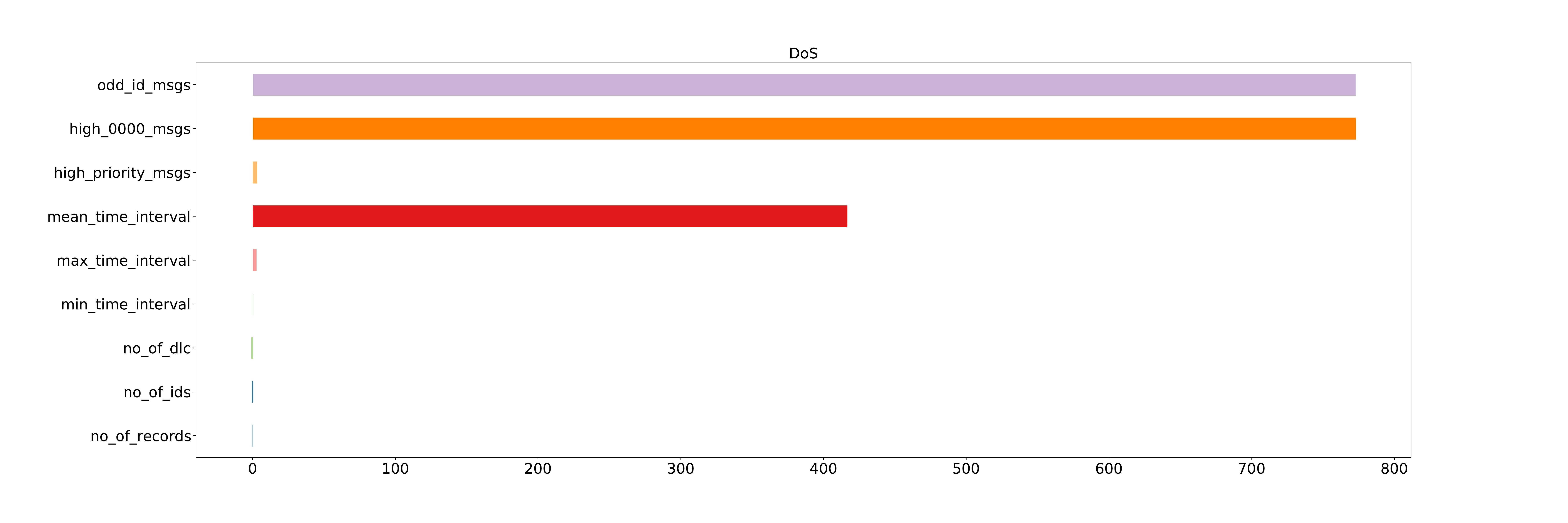}}\quad
\\
  \subfigure[Fuzzy]{\includegraphics[scale=0.23]{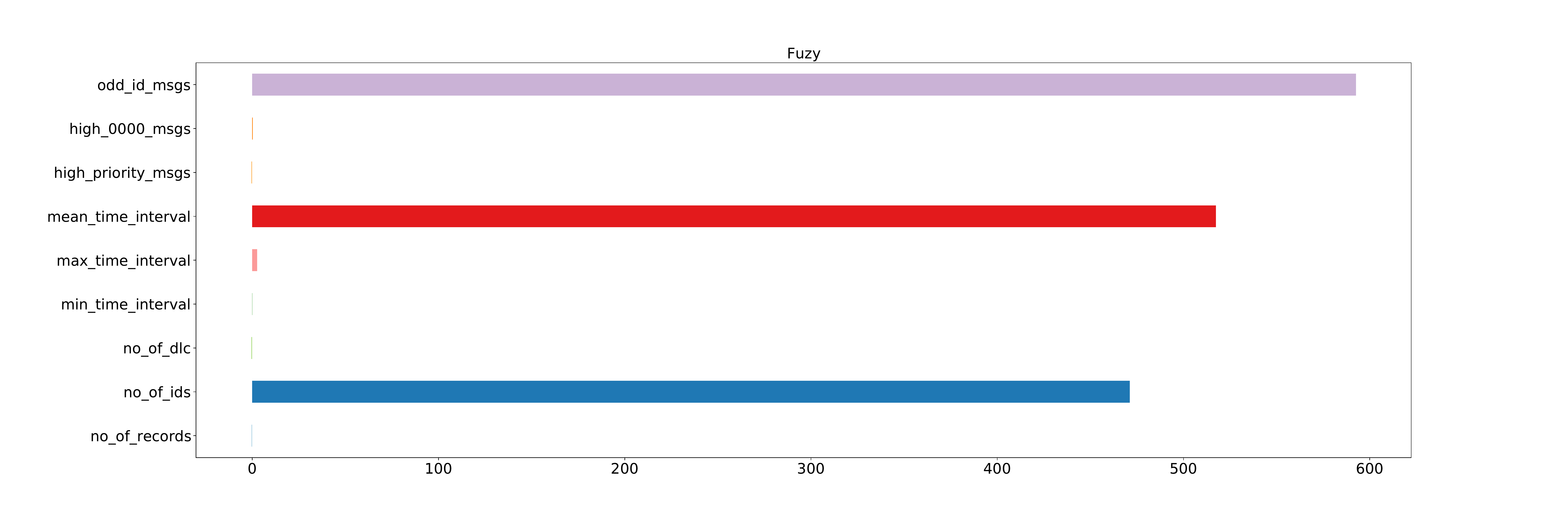}}\quad
\\
  \caption{Car Hacking Dataset: Outcomes generated using adversarial approach for DoS records and Fuzzy records \label{fig:global_explanation_carhacking} }
\end{figure*}

   We experimented with different millisecond time windows: 0.01, 0.02, 0.03, 0.04, 0.05.
    When looking at frames within the window, we observed some differences between normal and attack communication. 
    When the window size is too small, there are many windows without injected intrusion frames (During attack communication). 
    However, even without any injected frames, the window features of CAN frames are different due to the fact that these communication happens during attacks. 
    Thus, considering these windows as normal windows is inaccurate.  
   If the time window is too large, the number of records generated from window based feature extraction decreases. This results in less number of data records for training. 
    Out of the tested time windows, 0.03 and 0.04 milliseconds gave the best results. 
    The best-observed results were recorded in this section.
    Table \ref{table:ads_performance_comparison_carHacking} shows the detection performance of presented RX-ADS compared to recent state-of-the-art IDSs on Car hacking dataset: Histogram-based approach (HIDA) presented in \cite{9463874} and GIDS presented in \cite{8514157}.
    We calculate Accuracy, precision, recall, and F1 scores for comparison purposes with available literature. 

It can be seen that the anomaly detection rate of RX-ADS is higher for both DoS and Fuzzy intrusions compared to other approaches.
    As we discussed before, RX-ADS has an advantage over the HIDS approach as RX-ADS does not require labeled data for training. 
    Further, HIDS have implemented different variants of OCSVM-attack models for each intrusion, whereas RX-ADS only implements one model. 
    RX-ADS can use aggregated explanations for distinguishing DoS from Fuzzy intrusions.
    GIDS is similar to RX-ADS as they only train on normal data. 
    GIDS requires converting CAN data into image format for training \cite{8514157}. 
    Thus it is a complex and expensive pre-processing step compared to the simple window-based feature extraction used in RX-ADS.
    The major advantage of RX-ADS is the model interpretability, making domain experts verify model outcomes and debug and diagnose when necessary.

\begin{figure*}[t]
  \centering
  \includegraphics[scale=0.23]{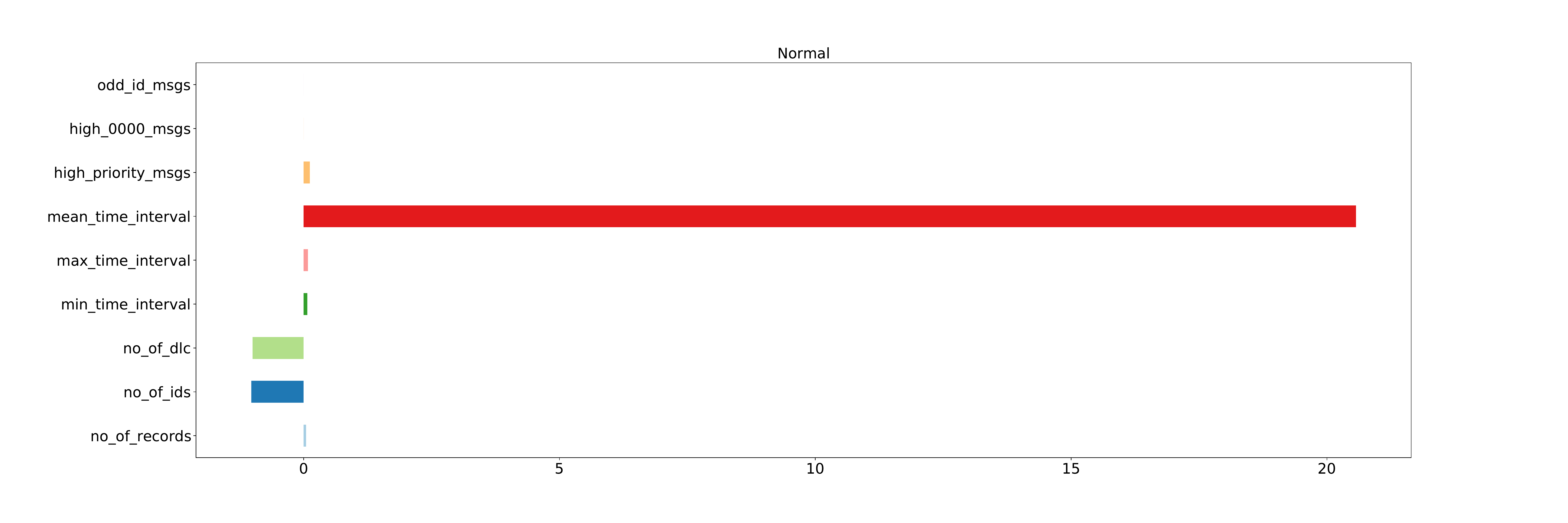}
  \caption{Car Hacking Dataset: Outcomes generated using adversarial approach for normal CAN frame communication during attacks (DoS) \label{fig:global_explanation_carhacking_Normal} }
\end{figure*}

    \subsubsection{Explanation generation}
    Figure \ref{fig:global_explanation_carhacking} shows the deviations calculated using adversarial approach for two types of abnormalities (DoS and Fuzzy).
    It can be seen that some features highly deviated during abnormal behaviors compared to baseline. 
    Further, there is a clear difference between the two abnormal behaviors. 
    Explanations for attack behaviors can be naturally interpreted in the following tabular format in Table \ref{table:carhacks_attack_vs_normal}.
    
    Compared to baseline, there are more frames with ID 0000 and odd IDs during DoS attacks. Further, the mean time interval between frames is higher during DoS attacks.
    Compared to baseline, a higher number of unique ID frames can be seen during Fuzzy attacks.

    
    We can also compare the behaviors between two attacks for distinguishing characteristics.
    During attack behaviors, both Fuzzy and DoS shows a higher number of odd ID frames (frame with IDs that have not been encountered during baseline behavior). However, during DoS, these are mainly are coming from frames with ID 0000, whereas in Fuzzy, these are not high priority frames. These are coming from random IDs which haven't encounter during baseline behavior. 
    Mean time interval between frames is higher for both compared to baseline. However, the Fuzzy attack shows the highest mean time interval than DoS.
    Number of unique IDs is very high during fuzzy attacks compared to DoS.
    
    Related literature on this dataset confirms the above-discussed behavior. For example, normal communication has a very low number of high-priority messages. In addition, the Fuzzy attacks result in CAN frames with random IDs which have not been encountered during baseline behaviors. During DoS, the number of high-priority messages with ID 0000 is higher compared to baseline and Fuzzy attacks. 
    Both attacks result in fewer frames within a time window. 
    This happens because, during attacks, it generates high-priority messages or spoofs random messages.
    These high-priority frames and other attack frames can have multiple effects, such as packet collisions and paralyzing the functions of a vehicle resulting in delays or even suspension of other messages \cite{8476919}.
    Therefore, the mean time interval between frames is higher compared to the baseline.
    These may be due to the CAN frame collisions and paralyzing the functions of a vehicle resulting in delays, or even suspension of other CAN messages \cite{8476919}.

    Figure \ref{fig:global_explanation_carhacking_Normal} shows the explanations generated for normal communication frames, which are in-between attack behaviors.
    Even though these samples do not have injected attack frames, the overall communication pattern of these windows was significantly different from the baseline. 
    Thus, the error threshold value was increased to detect these windows as normal windows. This can also be used as a similar filtering method to detect windows without injected frames within attacks (a similar filtering approach was proposed in HIDS). 
    These normal windows during attack communication are expected to be different from baseline communication as attack behaviors result in pre and post-effects in the systems, resulting in deviations from the baseline behavior. 
    These deviations seem to mainly result from higher mean time intervals between frames. Further, the number of unique IDs and DLC values seems low compared to the baseline. This matches domain experts' knowledge: attack communication results in a latency of CAN frames.
    These features need to be discussed with domain experts.



\section{Conclusion}
This work presented a ResNet Autoencoder based Explainable Anomaly Detection System (RX-ADS) for CAN bus communication data.
The ResNet Autoencoder model was used to learn the baseline/normal behavior from the CAN communication data. The reconstruction error threshold of ResNet Autoencoder was used to distinguish abnormal behaviors. 
The explanation generation method uses an adversarial sample generation approach for identifying the deviation of abnormal behavior from learned baseline behavior. 
This is achieved by finding the minimum modification required to covert abnormal samples to normal samples. 
These modifications are used to identify, visualize and explain the relevant feature behaviors for abnormalities. 
The approach was tested on two widely used benchmark CAN datasets released by the Hacking and Countermeasures Research Lab: OTIDS and Car Hacking.

RX-ADS detected abnormalities in the two benchmark CAN protocol datasets and showed comparable performance compared to the current work on OTIDS dataset while outperforming on Car Hacking dataset.
Further, the proposed approach is able to explain the abnormal behaviors of the intrusions matching the expert knowledge. 
The relevant features found by the presented approach helped with distinguishing between different abnormal behaviors.
Experimental results showed that the presented RX-ADS methodology provided insightful and satisfactory explanations for the selected datasets.
Further, to the best of our knowledge, this is the first attempt at developing an Explainable ADS for CAN protocol communication.

In future work, the proposed approach will be deployed in a real-world setting: INL EV Charging System setup.
Further, the proposed approach will be extended to add physical features for providing more holistic abnormal behavior detection in EV infrastructure.

\section*{Acknowledgements}
{
This work was supported in part by the Department of Energy through the U.S. DOE Idaho Operations Office under Contract  DE-AC07-05ID14517, and in part by the Commonwealth Cyber Initiative, an Investment in the Advancement of Cyber Research and Development, Innovation and Workforce Development (cyberinitiative.org).
}

\printbibliography[title={References}, heading=bibintoc]


\end{document}